\documentclass{article}

\usepackage[preprint]{neurips_2025}

\usepackage{amsmath}
\usepackage{graphicx}
\usepackage{algorithm}
\usepackage{algpseudocode}
\usepackage{multirow}

\usepackage[utf8]{inputenc} 
\usepackage[T1]{fontenc}    
\usepackage{hyperref}       
\usepackage{url}            
\usepackage{booktabs}       
\usepackage{amsfonts}       
\usepackage{nicefrac}       
\usepackage{microtype}      
\usepackage{xcolor}         

\title{Catastrophic Forgetting Mitigation Through Plateau Phase Activity Profiling
}

\author{%
  Idan Mashiach \\
  Department of Computer Science\\
  Bar-Ilan University\\
  \texttt{idan.mashiach@live.biu.ac.il} \\
  \And
  Oren Glickman \\
  Department of Computer Science \\
  Bar-Ilan University \\
  \texttt{
oren.glickman@biu.ac.il} \\
  \And
  Tom Tirer \\
  Faculty of Engineering \\
  Bar-Ilan University \\\texttt{tirer.tom@biu.ac.il} \\
}

\begin{document}

\maketitle

\begin{abstract}
Catastrophic forgetting in deep neural networks occurs when learning new tasks degrades performance on previously learned tasks due to knowledge overwriting. Among the approaches to mitigate this issue, regularization techniques aim to identify and constrain ``important'' parameters to preserve previous knowledge. In the highly nonconvex optimization landscape of deep learning, we propose a novel perspective: tracking parameters  during the final training plateau is more effective than monitoring them throughout the entire training process. We argue that parameters that exhibit higher activity (movement and variability) during this plateau reveal directions in the loss landscape that are relatively flat, making them suitable for adaptation to new tasks while preserving knowledge from previous ones. Our comprehensive experiments demonstrate that this approach achieves superior performance in balancing catastrophic forgetting mitigation with strong performance on newly learned tasks.
\end{abstract}

\section{Introduction}
Deep neural networks (DNNs) have demonstrated remarkable success across various tasks and domains \citep{krizhevsky2012imagenet,he2016deep,goodfellow2016deep}. However, when adapting pretrained DNNs to new tasks, they face a fundamental challenge dubbed ``catastrophic forgetting'' \citep{McCloskey1989,ratcliff1990connectionist}. This phenomenon, which is a well-known and significant issue in the field of continual learning \citep{thrun1995lifelong,parisi2019continual}, occurs when a network's performance on previously learned tasks significantly deteriorates as it learns new information, limiting its potential for continual learning and real-world deployment where adaptation to new tasks is essential \citep{DeLange2019,van2022three}.

Various approaches have been proposed to address catastrophic forgetting \citep{Kemker2018, ramasesh2021catastrophic}, with regularization methods being particularly attractive due to their minimal memory overhead and computational efficiency during inference. These methods aim to identify and constrain ``important'' parameters to preserve previous knowledge. A prominent approach in this category is Synaptic Intelligence (SI) \citep{zenke2017continual}, which tracks each parameter's contribution to loss reduction and total movement \textit{during the entire training}, penalizing changes to weights that significantly impacted previous tasks to keep them close to their values from those tasks. However, the optimization landscape of deep learning is highly nonconvex \citep{choromanska2015loss, kawaguchi2016deep, li2018visualizing, goodfellow2014qualitatively}, which suggests that the local geometry (as captured by gradients) in the early stage of training may not be indicative for the geometry at the final point.
This can result in overly restricting many parameters or failing to effectively rank their importance, as will be illustrated later in this paper.

In this paper, we propose a new method for catastrophic forgetting mitigation that measures parameter activity during the final training plateau phase, unlike existing methods that track parameter importance based on the entire training process. Our method introduces the Plateau Phase Activity Profile (PPAP), which captures more active parameters, through their higher and more varied movements, during this plateau phase to identify adaptable weights suitable for stronger updates while maintaining their role in the network.
Specifically,
we argue that these more active parameters during the final training plateau phase reveal directions in the loss landscape that are relatively flat, making them suitable for adaptation to new tasks while preserving knowledge from previous ones \cite{mirzadeh2020understanding, keskar2016large}.

We evaluate our method through two main experiments. First, we compare our approach against SI using their CIFAR10-CIFAR100 \cite{krizhevsky2009learning} experimental setup. Second, we conduct a comprehensive evaluation using a Leave One Class Out (LOCO) training scenario on CIFAR100's super-classes and fine-grained classes. Our results demonstrate that our method outperforms existing approaches, particularly in scenarios requiring a balance between catastrophic forgetting mitigation and last task performance.

\section{Problem Formulation}
\label{sec:problem_formulation}

Let us consider the continual learning setup, where a neural network with parameters $\theta \in \mathbb{R}^d$ is trained on a sequence of tasks $\{\mathcal{T}_1, \mathcal{T}_2, \dots, \mathcal{T}_n\}$, where each task $\mathcal{T}_t$ is being considered only after the model has been trained on all previous tasks $\{\mathcal{T}_1, \dots, \mathcal{T}_{t-1}\}$. Each task $\mathcal{T}_i$ consists of a dataset $\mathcal{D}_i = \{(x_j, y_j)\}_{j=1}^{m_i}$ where $x_j$ represents input datum and $y_j$ represents its corresponding label. When training on task $\mathcal{T}_{i+1}$, the network's parameters are updated to optimize performance on the new task, which can lead to deterioration in performance on previous tasks.

We seek a parameter update strategy that simultaneously optimizes performance on the current task while preserving performance on previous tasks. Formally, let $\mathcal{L}_i(\theta)$ represent the loss function of the network on task $\mathcal{T}_i$ with parameters $\theta$. Let $\theta_t$ denote the model parameters after training on tasks $\{\mathcal{T}_1, \dots, \mathcal{T}_t\}$. The goal can be expressed as finding parameters $\theta_{t+1}$ that solve:
\begin{equation}
\label{eq:min_optim_cata}
    \arg\min_{\theta_{t+1}} \mathcal{L}_{t+1}(\theta_{t+1}) \quad \text{subject to} \quad \mathcal{L}_j(\theta_{t+1}) \leq \mathcal{L}_j(\theta_j) + \epsilon_j \quad \forall j \leq t,
\end{equation}
where $\theta_{t+1}$ denotes the model parameters for the new task, $\mathcal{L}_{t+1}$ denotes the loss function for the new task, $\mathcal{L}_j$ denotes the loss function for previous task $j$, $\theta_j$ denotes the model parameters after training on task $j$, and $\epsilon_j$ is a small tolerance value that allows for a controlled amount of performance degradation on previous tasks.
Note that practical methods typically trade between such degradation and the performance on the new task using hyperparameter settings that do not have an explicit relation to $\epsilon_j$.

The success of any method addressing this optimization problem is typically measured through two key metrics: the accuracy in the current task (which reflects how well the method optimizes $\mathcal{L}_{t+1}$) and the accuracy in previous tasks (which reflects how well it satisfies the constraints). In practice, these metrics are often evaluated through validation accuracy rather than loss values, as accuracy provides a more interpretable measure of model performance. The trade-off between these metrics reflects the fundamental challenge of continual learning: maintaining a balance between learning new tasks and preserving knowledge from previous ones. This trade-off is particularly evident in our experimental setup, where we evaluated methods across different configurations and task sequences to assess their effectiveness.

\section{Related Work}

Research on catastrophic forgetting has produced several approaches, broadly categorized into different methodologies.

Architectural solutions \cite{mallya2018packnet, serra2018overcoming} modify the network structure to accommodate new knowledge while preserving old information. This includes approaches that allocate new resources or establish task-specific pathways, such as Progressive Neural Networks (PNN) \cite{rusu2016progressive} and Dynamically Expandable Networks (DEN) \cite{yoon2017den}, as well as Parameter-Efficient Fine-Tuning (PEFT) methods \cite{lester2021power, he2021parameter} like Low-Rank Adaptation (LoRA) \cite{hu2021lora} that freeze the base model and train only small adapters for new tasks. While these approaches are effective in preserving previous knowledge, they face scalability challenges and can limit performance on new tasks requiring significant adaptation.

Replay-based methods \cite{aljundi2019gradient, rebuffi2017icarl, prabhu2020gdumb}, such as Experience Replay (ER) \cite{chaudhry2018er} and Deep Generative Replay (DGR) \cite{shin2017dgr}, maintain access to previous task information through storage or generation of past experiences. These methods rely on previous task data during training on new tasks to prevent forgetting. While showing strong performance, they require significant memory or computational resources, making them less suitable for long-term continual learning scenarios.

Regularization methods \cite{zenke2017continual,kirkpatrick2017ewc,li2017lwf,aljundi2018memory,lopez2017gradient,farajtabar2020orthogonal,chaudhry2018riemannian, nguyen2018variational} form the foundation for our work. These approaches are particularly attractive due to their minimal memory overhead and ability to maintain the original network architecture without requiring additional components during training or inference. Knowledge distillation approaches preserve previous task performance by maintaining similar outputs to the original model, effectively regularizing the network's behavior. Examples include Learning without Forgetting (LWF) \cite{li2017lwf}, which requires storing activation patterns from the original network, and Memory Aware Synapses (MAS) \cite{aljundi2018memory}, which estimates parameter importance by measuring output sensitivity to parameter changes. Gradient-based methods manipulate gradient updates to prevent catastrophic forgetting. Examples include Gradient Episodic Memory (GEM) \cite{lopez2017gradient}, which projects new gradients to avoid increasing loss on stored examples, and Orthogonal Gradient Descent (OGD) \cite{farajtabar2020orthogonal}, which projects gradients to minimize interference with previous knowledge. 

A specific class of regularization strategies is
parameter-based methods, which aim to directly identify weights deemed important for the trained task. During training on subsequent tasks, these parameters are either protected from change or penalized when modified, while the remaining weights are updated. Notable examples include Elastic Weight Consolidation (EWC) \cite{kirkpatrick2017ewc} and Synaptic Intelligence (SI) \cite{zenke2017continual}, which we use as baselines. SI, in particular, estimates parameter importance by accumulating gradients and their impact throughout training, assigning higher importance to parameters that significantly affect the loss. While effective in simple scenarios, this approach may not accurately reflect parameter importance in complex loss landscapes, as it relies on gradient magnitudes during early active learning phases where the loss is rapidly decreasing.

Our method differs from these approaches by focusing on parameter activity during the loss plateau rather than tracking their gradients throughout the entire learning phase. This provides a measure of parameter \textit{flexibility}---ability to be adapted without significantly affecting the loss---which fits complex loss landscapes where gradient magnitudes during early learning may not accurately reflect the parameter significance.

\section{Plateau Phase Activity Profile (PPAP)}
\label{sec:ppap}

We propose a novel method that mitigates catastrophic forgetting by learning a weight-level profile during the final plateau phase of task $t$. This profile, which we term the Plateau Phase Activity Profile (PPAP), assigns a ``flexibility score'' to each parameter based on their behavior during the final period where the model's learning was insignificant. The key insight behind focusing on this phase is that parameters exhibiting large activity indicate a wider range of flexibility and adaptability without affecting the loss, making them more suitable for stronger updates in subsequent tasks.

Our method relies on optimizers with a momentum component, such as SGD with momentum or Adam, which are standard choices.
These optimizers mitigate 
the effect of unstable gradients on the parameter update, and thus smooth the optimization trajectory.
This smoothness is essential for our method, as it allows us to accurately measure parameter activity and reliably identify which parameters exhibit higher activity levels during the final plateau phase.

\begin{figure}[t]
    \centering
    \includegraphics[width=0.8\textwidth]{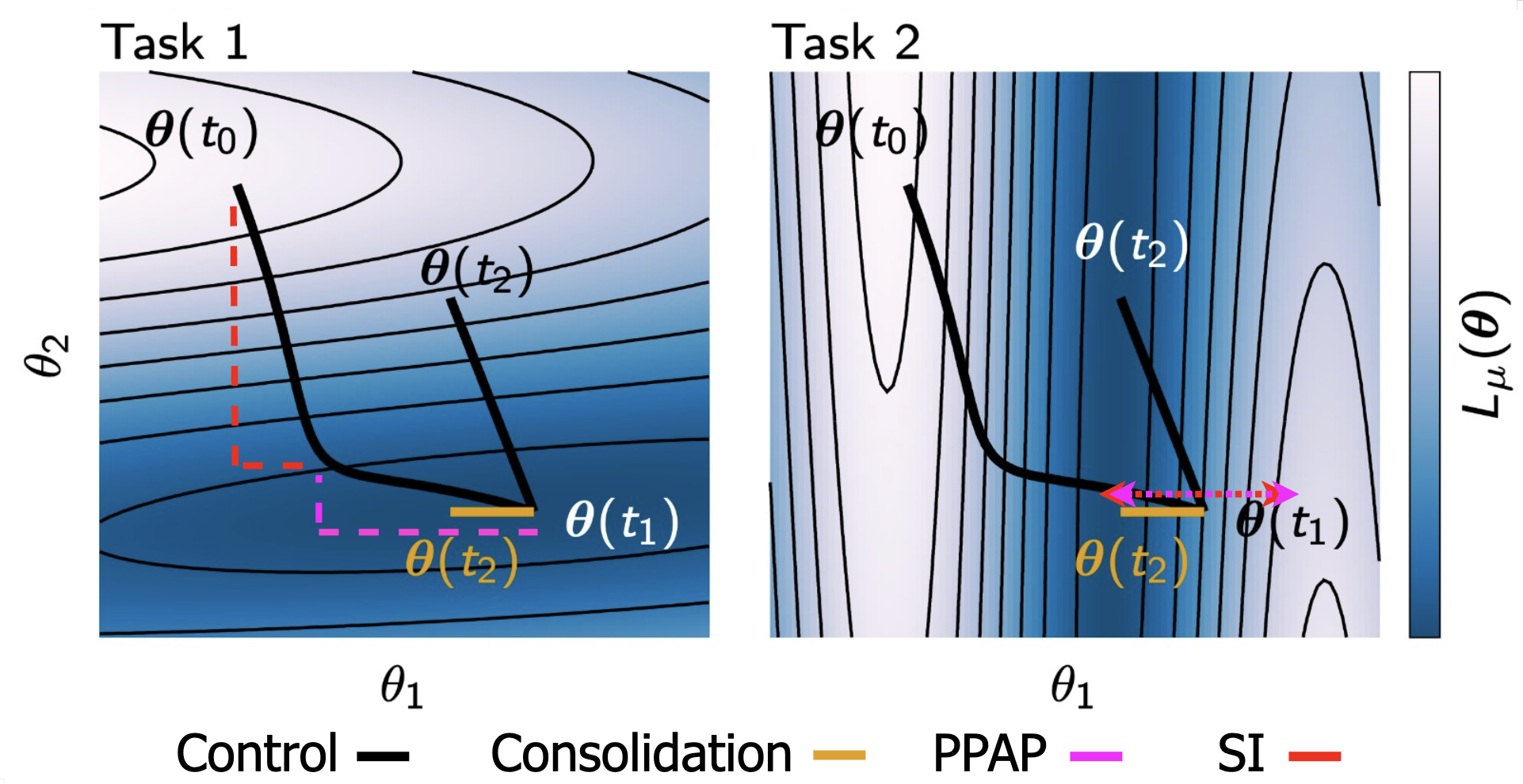}
    \caption{Schematic illustration of parameter space trajectories in two tasks (taken from \citep{zenke2017continual}). The heat map represents loss function $L_\mu$ values, with solid lines showing parameter trajectories. Gradient descent during Task 1's training moves from $\theta(t_0)$ to $\theta(t_1)$. Plain  Gradient descent during Task 2's training moves from $\theta(t_1)$ to $\theta(t_2)$, increasing Task 1's loss. The orange point $\theta(t_2)$ shows an alternate solution with low loss for both tasks. In Task 1, red dotted lines show the dominant part of SI's parameter importance tracking, while pink dotted lines show the Plateau Phase Activity Profile (PPAP)'s tracking. Red and pink dotted arrows indicate gradient scaling direction, where both methods prioritize $\theta_1$ direction to achieve low loss for both tasks. }
    \label{fig:parameter_trajectories}
\end{figure}

\begin{figure}[t]
    \centering
    \includegraphics[width=0.8\textwidth]{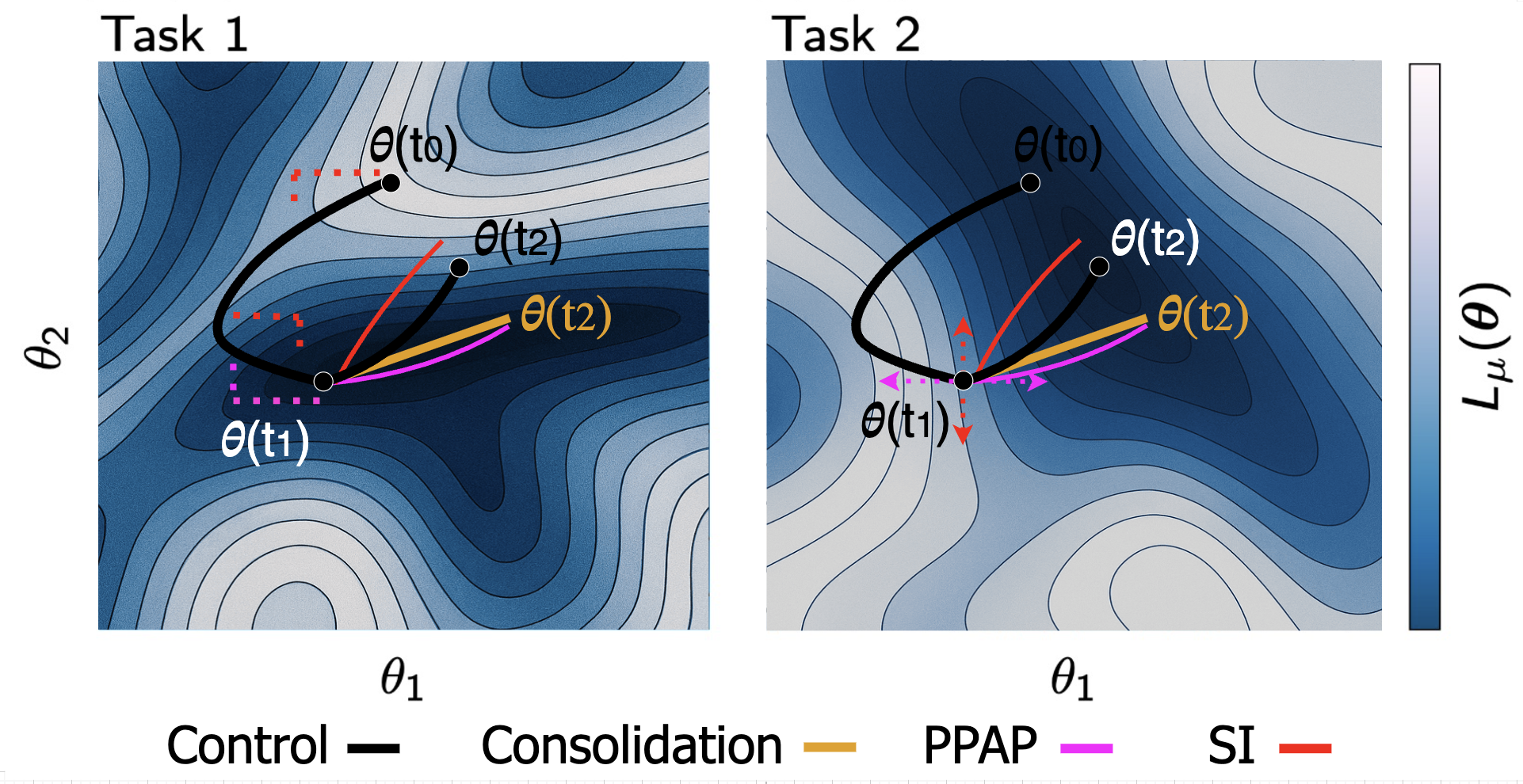}
    \caption{Schematic illustration of a complex parameter space trajectories demonstrating the limitations of SI in catastrophic forgetting mitigation. Similar to Figure \ref{fig:parameter_trajectories}, the heat map shows loss function $L_\mu$ values, with solid lines indicating parameter trajectories. In this complex scenario, SI assigns a high score to $\theta_1$ and a low score to $\theta_2$ (red dotted lines) because $\theta_1$ exhibited higher gradients during Task 1's rapid loss descent. This leads SI to scale gradients in the $\theta_2$ direction during Task 2 (red dotted arrows), resulting in a suboptimal solution (solid red line). In contrast, our PPAP assigns a high score to $\theta_1$ and a low score to $\theta_2$ (pink dotted lines) because $\theta_1$ is more active during the loss plateau, scaling gradients in the $\theta_1$ direction (pink dotted arrows) and leading to the alternate solution (solid pink line) that achieves low loss for both tasks.}
    \label{fig:complex_parameter_trajectories}
\end{figure}

\subsection{Core Idea and Motivation}

Based on the standard continual learning objective \eqref{eq:min_optim_cata}, we measure each parameter $\theta_w$'s activity during the final loss plateau by combining its overall movement and variability. Specifically, we look how much the parameter's movement contributes to progress along the loss surface based on aggregating the product of parameter updates $\Delta\theta_w$ and their corresponding gradients $\frac{\partial \mathcal{L}}{\partial \theta_w}$ over the plateau period $[T_0,T]$, measuring both the total magnitude of these changes and their variation over time.

Parameters that show higher flexibility scores during this plateau phase demonstrate an important property: they can be modified more substantially without significantly affecting the loss $\mathcal{L}_j$ of previous tasks. This means that for these parameters, there exists a wider range of possible parameter change $\delta \theta_i$ that maintain the loss change $\Delta \mathcal{L}_j = \mathcal{L}_j(\theta_{t+1}) - \mathcal{L}_j(\theta_j)$ within the acceptable bound $\epsilon$. In other words, these parameters have more freedom to change while still satisfying the loss minimization constraint $\mathcal{L}_j(\theta_{t+1}) \leq \mathcal{L}_j(\theta_t) + \epsilon_j$ for previous tasks.

We turn now to highlight the difference between the mechanisms of SI and the proposed PPAP through schematic illustrations. 
Figure \ref{fig:parameter_trajectories}, taken from the SI paper \citep{zenke2017continual}, illustrates a simplified scenario where both SI and our PPAP successfully mitigate catastrophic forgetting, though through different mechanisms. During Task 1, SI assigns a high ``importance score'' to $\theta_2$ and a low score to $\theta_1$ because $\theta_1$ exhibited greater gradients that significantly affected the loss. In contrast, our method assigns a high ``flexibility score'' to $\theta_1$ and a low score to $\theta_2$ because during the final loss plateau--where learning was minimal--$\theta_1$ exhibited more activity than $\theta_2$. During Task 2, both approaches ultimately constrain $\theta_2$ while allowing more training freedom for $\theta_1$, though for different reasons: SI constrains $\theta_2$ more due to its high importance score (along training in Task 1), while our method enables more training for $\theta_1$ due to its higher flexibility score (activity in the plateau phase in Task 1). However, this illustration lacks the complexity that reveals SI's limitations, which our PPAP overcomes.

In a more complex scenario, depicted in Figure \ref{fig:complex_parameter_trajectories}, the different approaches lead to significantly different outcomes. While SI's focus on gradients during rapid loss descent causes it to weigh more the $\theta_2$ dimension, leading to a suboptimal minimum (solid red line), our method's focus on activity during the plateau phase, enables it to weigh more the $\theta_1$ dimension, resulting in a better minimum for catastrophic forgetting mitigation (solid pink line). This demonstrates that learning activity during plateau periods is crucial, as relying on gradients that most affect the loss throughout the entire training process can lead to suboptimal convergence in complex loss spaces.

Note that these illustrations (motivated by the one in \citep{zenke2017continual}) depict idealized gradient descent trajectories with infinitesimal step sizes (commonly referred to as ``gradient flow''), rather than the practical stochastic gradient descent (SGD) used in real training. In practice, SGD/Adam introduce noise in the training trajectory, and when reaching the final minimum area, the network jitters around it as long as the learning rate remains constant. These stochastic behaviors are actually beneficial for our method, as they introduce additional dynamics that enable more accurate identification of adaptable parameters for catastrophic forgetting mitigation.

\subsection{Learning the Plateau Phase Activity Profile}

For efficient computation of the PPAP, we use an incremental approach that enables low memory usage and parallel operations. This design maintains a constant memory footprint regardless of the number of training steps.
For each training step $i$ in task $t$, we compute the loss difference $\Delta \mathcal{L}^i$ calculated as:
\begin{equation}
    \Delta \mathcal{L}^i = \mathcal{L}(\theta_{i+1}; \mathcal{B}_i) - \mathcal{L}(\theta_i; \mathcal{B}_i),
\end{equation}
where $\mathcal{L}$ is the loss function, $\mathcal{B}_i$ is a training mini-batch, $\theta_i$ are the model parameters before the optimizer step $i$, and $\theta_{i+1}$ are the model parameters after optimizer step $i$.

We pass $\Delta \mathcal{L}^i$ into a Gaussian function to properly scale parameter activity during plateau phases, where the loss is stable and parameter updates should be weighted more heavily:
\begin{equation}
    f(\Delta \mathcal{L}^i) = e^{-k \cdot (\Delta \mathcal{L}^i)^2}.
\end{equation}
The hyperparameter $k$ controls the width of the Gaussian: larger values result in a narrower Gaussian that is more sensitive to small changes, while smaller values result in a wider Gaussian that is less sensitive to changes. The function reaches its maximum value of 1 when $\Delta \mathcal{L}^i = 0$ and approaches 0 as the loss difference increases.

For a single parameter $w$ in $\theta$,
let $\Delta\theta_w^i = \theta_w^{i+1} - \theta_w^i$ be the parameter update at step $i$, and $\frac{\partial \mathcal{L}^i}{\partial \theta_w}$ the corresponding partial derivative of $\mathcal{L}(\theta_i; \mathcal{B}_i)$. For each parameter $w$, we compute the parameter's plateau phase activity at iteration $i$, denoted by $A_w^i$, as:
\begin{equation}
    A_w^i = \Delta\theta_w^i \cdot \frac{\partial \mathcal{L}^i}{\partial \theta_w} \cdot f(\Delta \mathcal{L}^i)
\end{equation}
Using these $\{A_w^i\}$, we compute two accumulating measurements for each weight:
\begin{enumerate}
    \item The sum of absolute $A_w^i$:
    \begin{equation}
        S_w = \sum_{i=1}^{N} |A_w^i|,
    \end{equation}
    where $N$ is the total number of training steps.
    \item The standard deviation of $A_w^i$:
    \begin{equation}
        \sigma_w = \sqrt{\frac{1}{N}\sum_{i=1}^N (A_w^i - \mu_w)^2},
    \end{equation}
    where $\mu_w$ is the mean of $A_w^i$ for weight $w$. For implementation details of the online computation, see Appendix \ref{sec:online_std}.
\end{enumerate}

During training, whenever the loss $\Delta \mathcal{L}^i \geq \sqrt{\frac{1}{2k}}$ (the standard deviation of our Gaussian function), we reduce the weight of accumulated measurements by multiplying both measurements by $f(\Delta \mathcal{L}^i)$. This reduction is crucial as a significant $\Delta \mathcal{L}^i$ indicates that the network is entering a new learning phase, making previous plateau measurements less relevant.

At the end of training, we merge the two measurements into the PPAP. We first normalize each feature using min-max normalization to the range [0,1]:
\begin{equation}
    S_w^{norm} = \frac{S_w - \min(S)}{\max(S) - \min(S)} \quad \text{and} \quad \sigma_w^{norm} = \frac{\sigma_w - \min(\sigma)}{\max(\sigma) - \min(\sigma)},
\end{equation}
where $\min(\cdot)$ and $\max(\cdot)$ are computed across all entries in the respective data structure, e.g., $\min(S)=\min_{w'}(S_{w'})$.

We use min-max normalization because it preserves extreme values that contain important information about weight behavior, unlike other normalization methods that tend to compress or eliminate these signals. The normalized measurements are then merged according to:
\begin{equation}
    P_w^{pre} = S_w^{norm} \cdot \sigma_w^{norm}
\end{equation}
Finally, we normalize the merged values to obtain the PPAP:
\begin{equation}
    P_w = \frac{P_w^{pre} - \min(P^{pre})}{\max(P^{pre}) - \min(P^{pre})}.
\end{equation}
This profile represents how much each weight can be modified in subsequent tasks while preserving the knowledge learned in task $t$. Weights with scores closer to 1 are more flexible and can be trained more strongly in subsequent tasks, while weights with scores closer to 0 are more stable and should be trained more conservatively to preserve the knowledge learned in task $t$.

For a detailed pseudo-code of the PPAP learning algorithm, see Appendix \ref{sec:computing_pfp_algorithm}.

\subsection{Using the Plateau Phase Activity Profile}

The PPAP is integrated into the optimization process of task $t+1$ through a specialized weight update mechanism. Unlike traditional regularization methods that add components to the loss function, we implement this through custom optimizer extensions that preserve the original optimization mechanisms while adding hooks for weight update modification. This design enables direct modification of weight updates based on the learned scores.\footnote{This prevents different utilization of PPAP for different optimizers that can result from loss modification. For example, weight decay is equivalent to squared $\ell_2$-norm regularization in SGD but not in Adam \citep{loshchilov2017decoupled}.}

Let $P$ denote the PPAP score of all parameters (has the dimension of $\theta$).
Denote by $\Delta \theta^i$ the update step computed by the optimizer at the $i$-th training iteration of task $t+1$ (e.g., for Adam, this is the first moment scaled by the second moment).
We modify this update as follows:
\begin{equation}
    \Delta \theta^i_{modified} = (r \cdot \Delta \theta^i) + ((1 - r) \cdot \Delta \theta^i \odot P),
\end{equation}
where $\odot$ denotes elements-wise product, and
$r$ is a hyperparameter that controls the balance between regular updates and updates modulated by the profile. The modified update is then used by the optimizer in its weight update phase. This formulation ensures that a fraction $r$ of the update follows the standard optimization process, while the remaining fraction $(1-r)$ is scaled by the weight's corresponding PPAP score.

\section{Experiments}
\label{sec:experiments}
We evaluate the effectiveness of PPAP in mitigating catastrophic forgetting across two experimental setups: a sequential CIFAR10-CIFAR100 benchmark and a Leave-One-Class-Out (LOCO) setup on CIFAR100. In both cases, after each task, the performance is evaluated by the accuracy on a validation set (unseen during training). We report results for both the current task (adaptation) and previously learned tasks (retention).

\subsection{CIFAR10-CIFAR100}
\subsubsection{CIFAR10-CIFAR100 Setup}
\label{sec:cifar-10-cifar100-setup}

We first compare PPAP with Synaptic Intelligence (SI) \citep{zenke2017continual} using the evaluation setup described in their paper, approximating details when unspecified. Our implementation faithfully reproduces their results, providing a reliable basis for comparison.

The model is a CNN with task-specific output heads. From Task 2 onward, we apply the PPAP profile computed from the previous tasks. The training procedure follows:
\begin{itemize}
    \item Task 1: Train on the full CIFAR-10 dataset.
    \item Tasks 2-6: Sequentially train on 5 additional tasks, each corresponding to 10 consecutive classes from the CIFAR-100 dataset.
\end{itemize}
For more details, see Appendix \ref{sec:cifar10_cifar100_setup_details}.

\subsubsection{CIFAR10-CIFAR100 Results}
\begin{figure}[t]
    \centering
    \includegraphics[width=\textwidth]{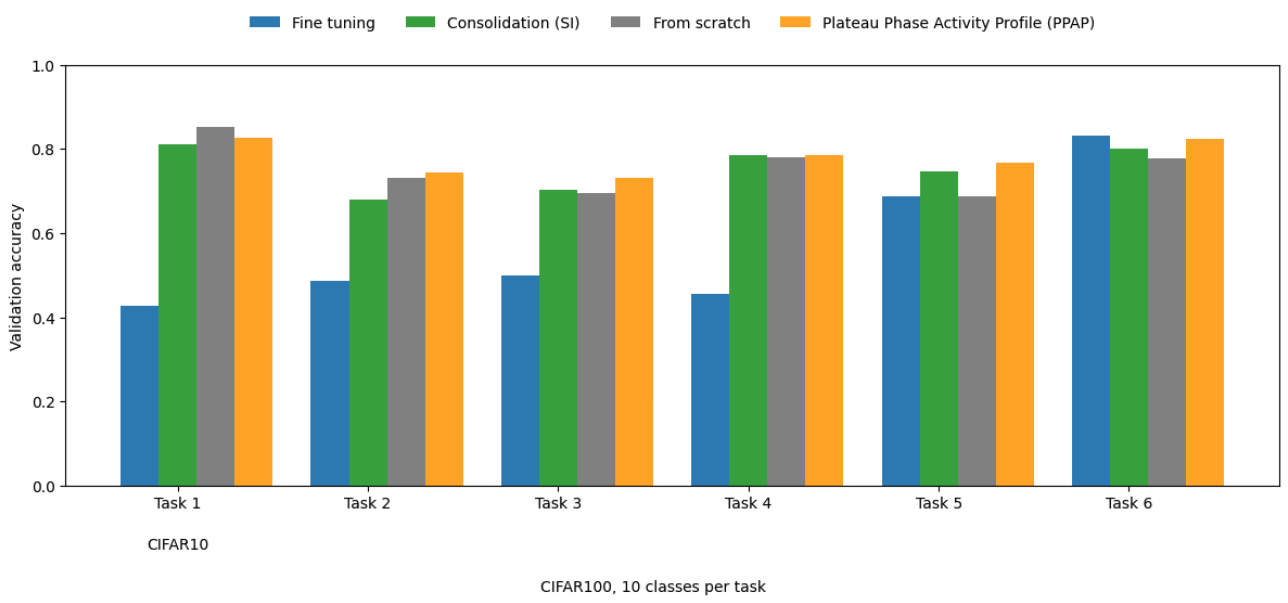}
    \caption{Comparison of different training approaches on CIFAR10-CIFAR100 tasks. Gray bars represent "From scratch" performance (training a new model for each task), blue bars show "Fine tuning" results (sequential training without mitigation), green bars indicate SI's "Consolidation" results, and orange bars display our PPAP performance. Each approach, except "From scratch", is evaluated across six tasks, with accuracy measured on each task's validation data. See Table \ref{tab:cifar10_cifar100_results} in Appendix \ref{app:results} for detailed numerical results.}
    \label{fig:cifar10-cifar100-results}
\end{figure}

The results for the setup are presented in Figure \ref{fig:cifar10-cifar100-results}.
Following the evaluation methodology of SI, we compare our method's performance using their established metrics. The evaluation uses three key metrics for each task: "From scratch" (gray bar) represents the accuracy obtained by training a new model specifically on that task's data, serving as a baseline for optimal performance; "Fine tuning" (blue bar) shows the accuracy after sequential training on all tasks, i.e., after Task 6, revealing the extent of catastrophic forgetting; and "Consolidation" (green bar) demonstrates the performance with SI's mitigation method applied. We added our PPAP evaluation (orange bar) using the same sequential training approach but with our mitigation method.

As reported in the SI paper, their results show that networks trained with consolidation maintain consistent validation accuracy across all tasks, while those without consolidation show a clear decline in performance on older tasks. The consolidation approach consistently outperforms the non-consolidated version on all tasks except the last one. The comparison between consolidated and single-task networks shows that their performance is approximately equal across tasks, with slight variations in favor of single-task networks at the beginning and consolidation at the end.

Our results demonstrate that PPAP method effectively mitigates catastrophic forgetting across all tasks. It consistently achieves equal or better accuracy than SI over all tasks. This superior performance is evident both in mitigating catastrophic forgetting (tasks 1-5) and in the final task performance (task 6), highlighting our method's effectiveness in both aspects of continual learning.

\subsection{LOCO CIFAR100}
\subsubsection{LOCO CIFAR100 Setup}
\label{sec:loco-cifar100-setup}

To evaluate performance more extensively, we employ a ResNet18 \cite{he2016deep} architecture in a Leave One Class Out (LOCO) training scenario on CIFAR100, where we compare our method with SI and EWC \citep{kirkpatrick2017ewc} to demonstrate its effectiveness in a more complex setup. In this scenario, we systematically leave out one of the 20 superclasses during pretraining and then finetune on its 5 fine-grained classes.

In this setup, we focus on two consecutive tasks: pretraining (task $t$) and finetuning (task $t+1$). To ensure robust evaluation, we perform an extensive analysis across different training configurations:
\begin{itemize}
    \item For each superclass index $i$ (where $i \in \{0, \ldots, 19\}$):
    \begin{itemize}
        \item Pretraining (task $t$): Train on all superclasses except $i$, where each image is labeled with its superclass.
        \item Finetuning (task $t+1$): Train on the 5 fine-grained classes of superclass $i$.
    \end{itemize}
    \item For each of these 20 superclass combinations, we evaluate four epoch configurations: (20, 20), (100, 20), (100, 100), (600, 100), where the first number represents pretraining epochs and the second represents finetuning epochs.
\end{itemize}

Final metrics are computed by averaging the validation accuracies across all 20 superclass combinations for each epoch configuration, separately for both pretraining and finetuning tasks.

Unlike the CIFAR10-CIFAR100 setup which uses separate heads for each task, we employ a linear probing approach after finetuning.\footnote{This is motivated by recent works \citep{kirichenko2023last,qiu2023simple} that show that even if the classifier's accuracy is low, its feature mapping (deepest representations) may capture/preserve valuable patterns.} That is, we freeze the network except for the last classification layer, and train only this  layer on the pretraining data. This ensures the classification layer is properly optimized for the current network state, allowing us to measure the quality of the model's \textit{feature mapping} (deepest representations) on the pretraining data when evaluating the catastrophic forgetting mitigation approach.
All linear probing trainings were trained for 600 epochs to ensure optimal performance of the last layer classifier. For more details, see Appendix \ref{sec:loco_cifar100_setup_details}.

\subsubsection{LOCO CIFAR100 Results}

\begin{figure}[t]
    \centering
    \includegraphics[width=0.3\textwidth]{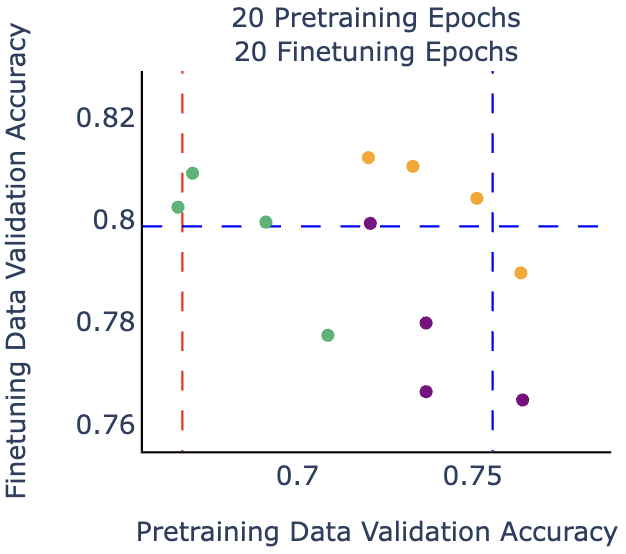}
    \hspace{0.02\textwidth} 
    \includegraphics[width=0.3\textwidth]{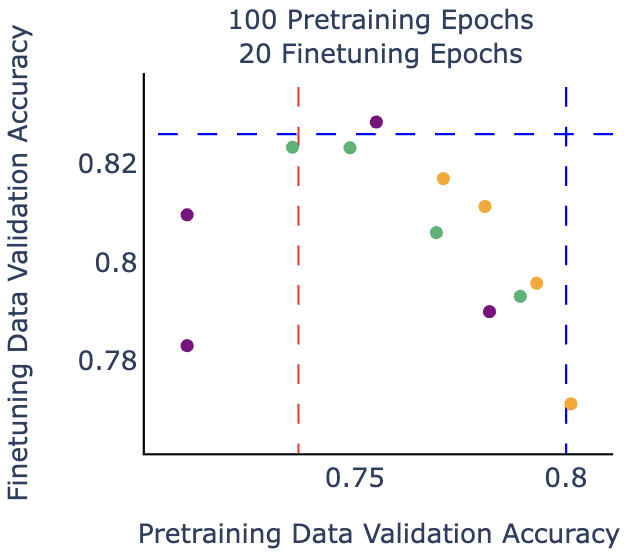}
    \hspace{0.02\textwidth}
    \\[1em] 
    \includegraphics[width=0.3\textwidth]{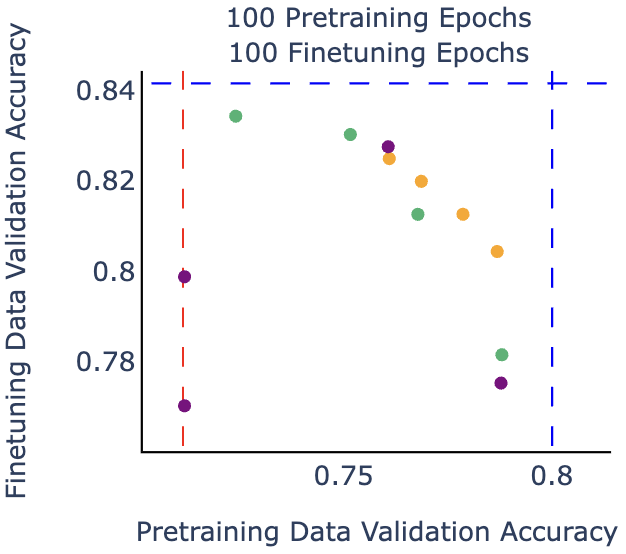}
    \hspace{0.02\textwidth}
    \includegraphics[width=0.3\textwidth]{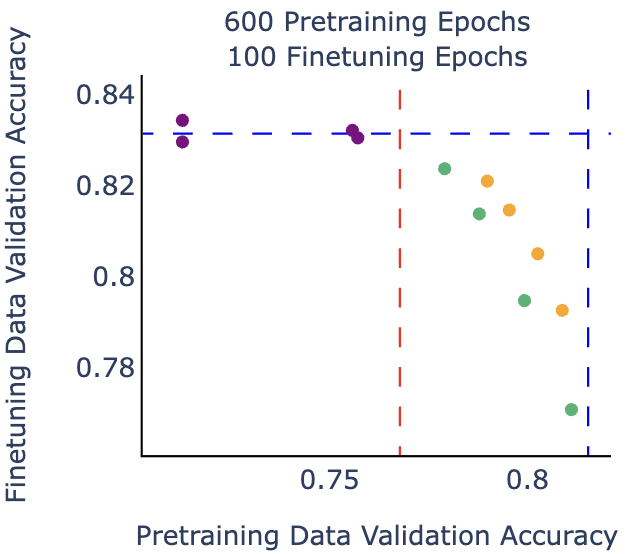}
    
    \caption{LOCO CIFAR100 Results: Comparison of different methods using Cartesian coordinates, where the horizontal axis shows pretraining accuracy (19 superclasses) and the vertical axis shows finetuning accuracy (5 fine-grained classes). Each dot represents average performance across 20 choices of superclass  (details in the text). Dashed lines indicate references: rightmost blue (pretraining target), leftmost red (baseline degradation), and upper blue (finetuning target). Purple dots: EWC, Green Dots: SI, Orange Dots: PPAP. The different points of the same color represent different regularization strength hyperparameter. See Table \ref{tab:loco_cifar100_results} in Appendix \ref{app:results} for detailed numerical results.}
    \label{fig:loco-results}
\end{figure}

In the CIFAR10-CIFAR100 setup, the evaluation focused on the final model's performance after multiple consecutive task trainings, presenting the results as bar charts showing only the last model's accuracies. In contrast, our LOCO CIFAR100 setup involves only two consecutive tasks (pretraining and finetuning), allowing us to represent the results more comprehensively using Cartesian coordinates.

The results of our LOCO CIFAR100 setup are presented in Figure \ref{fig:loco-results}.
We present a separate Cartesian coordinate chart for each epoch configuration, with three dashed  lines providing meaningful context. The rightmost dashed line shows the pretraining data accuracy at the end of pretraining, serving as an ``upper bound'' or ``target'' for pretraining performance (potentially, it can be passed since the second task's data may be useful for the first task). The leftmost dashed line represents the pretraining data accuracy immediately after finetuning, indicating the baseline degradation caused by finetuning -- any performance below this value suggests the mitigation technique is not effective. The upper dashed line shows the finetuning data accuracy immediately after finetuning, serving as the ``upper bound''/``target'' for finetuning performance.

The ideal performance would be in the upper-right corner of the chart, indicating both strong catastrophic forgetting mitigation and minimal sacrifice in finetuning performance.
Yet, due to the natural tradeoff between them we explore different values of regularization strength hyperparameter for each of the methods (values of $r$ for PPAP) and explore their frontiers.

From Figure \ref{fig:loco-results}, we see that for the case of small number of epochs our approach dominates SI and EWC.
In the other cases, 
none of the methods has has perfect performance in both dimensions (close to the intersection of the ``upper bounds''). In such cases, the choice between methods depends on the client's preference for balancing catastrophic forgetting mitigation against finetuning performance. For clients seeking a balanced score, the Euclidean distance from the origin (0,0) may serve as an effective metric, as it equally weights both dimensions - a higher distance indicates better overall performance in both catastrophic forgetting mitigation and finetuning accuracy. Our method consistently achieves higher Euclidean distances than both SI and EWC approaches across most configurations reflecting the best trade-off between catastrophic forgetting mitigation and new task performance.

\section{Conclusion}

In this paper, we proposed a new method for catastrophic forgetting mitigation, named the Plateau Phase Activity Profile (PPAP), which is based on tracking parameters with high activity during the training's final plateau phase, associated with directions around the minimizer where the loss landscape is relatively flat, making them suitable for adaptation to new tasks while preserving knowledge from previous ones.
We systematically evaluated our approach and showed its advantages over popular alternatives.

The current PPAP implementation demonstrates effectiveness in mitigating catastrophic forgetting while maintaining strong finetuning performance. Nevertheless, several promising directions may enhance its performance. First, one may explore layer-wise normalization to better capture the relative importance of different network layers. 
Second, our PPAP approach may benefit from optimizers that aim at reaching flatter minima \citep{foret2021sharpnessaware,refael2025lorenza}, since it will increase the number of flexible parameters that can be safely adapted to new tasks. Lastly, the relation of PPAP with transfer learning performance may also inform better pretraining strategies and architectural innovations for continual learning.

\section*{Acknowledgment}
TT is supported by the Israel Science Foundation (No. 1940/23) and MOST (No. 0007091) grants.

We would like to thank SwarmOne for providing free and unlimited access to their innovative Autonomous AI Infrastructure Platform, which enabled us to conduct the extensive number of training runs required for our research goals.



\bibliographystyle{unsrt}
\label{sec:references}
\bibliography{refs}


\newpage

\appendix

\section{Online Standard Deviation Computation}
\label{sec:online_std}

The standard deviation is computed using an online algorithm with the following update rules. We initialize $\mu^0 = 0$ and $SSD^0 = 0$. 
For every step $i$, we calculate the running mean:
    \begin{equation}
        \mu^{i+1} = \mu^i + \frac{x^i - \mu^i}{i+1}
    \end{equation}
where $x^i$ is the value at step $i$. 

Then, we compute the running sum of squared differences using:
    \begin{equation}
        SSD^{i+1} = SSD^i + (x^i - \mu^i) \cdot (x^i - \mu^{i+1})
    \end{equation}
At the end of training, the standard deviation is computed as:
    \begin{equation}
        \sigma = \sqrt{\frac{SSD^N}{N}}
    \end{equation}

To reduce the weight of accumulated measurements in an online standard deviation calculation, we adjust both the step counter $N$ and the running sum of squared differences $SSD$. We multiply $N$ by a reduction factor $r$ (rounded up to maintain integer steps) and $SSD$ by $r^2$. This ensures the statistical properties of the online calculation are properly maintained while reducing the influence of past measurements.

\section{Computing the Plateau Phase Activity Profile (PPAP) Algorithm}
\label{sec:computing_pfp_algorithm}
The full PPAP algorithm is presented in Algorithm \ref{alg:pfp}.

\begin{algorithm}
\caption{Computing the Plateau Phase Activity Profile (PPAP)}
\label{alg:pfp}
\begin{algorithmic}[1]
\Require Training data $\mathcal{D} = \{(x_i, y_i)\}$, model parameters $\theta$, $k$ (Gaussian width)
\Ensure PPAP $P_w$ for each weight $w$

\State Initialize $S_w = 0$, $\mu_w = 0$, $SSD_w = 0$ for all weights $w$
\State $\sigma = \sqrt{\frac{1}{2k}}$ \Comment{Standard deviation of Gaussian}

\For{each training step $i$}
    \State \Comment{Regular training step}
    \State Compute current loss $\mathcal{L}_i = \mathcal{L}(x_i, y_i, \theta_i)$
    \State Compute gradients $\frac{\partial \mathcal{L}}{\partial \theta_w}$ for all weights $w$
    \State Apply optimizer step to update model parameters $\theta_{i+1}$
    
    \State \Comment{PPAP computation}
    \State Compute new loss $\mathcal{L}_{i+1} = \mathcal{L}(x_i, y_i, \theta_{i+1})$ \Comment{Forward pass only, same data}
    \State Compute $\Delta \mathcal{L}^i = \mathcal{L}_{i+1} - \mathcal{L}_i$
    \State Compute scaling factor $f(\Delta \mathcal{L}^i) = e^{-k \cdot (\Delta \mathcal{L}^i)^2}$
    
    \If{$\Delta \mathcal{L}^i > \sigma$}
        \State $S_w = S_w \cdot f(\Delta \mathcal{L}^i)$ for all weights $w$
        \State $N_w = \lceil N_w \cdot f(\Delta \mathcal{L}^i) \rceil$ for all weights $w$
        \State $SSD_w = SSD_w \cdot f(\Delta \mathcal{L}^i)^2$ for all weights $w$
    \EndIf
    
    \For{each weight $w$}
        \State Compute parameter update: $\Delta\theta_w^i = \theta_w^{i+1} - \theta_w^i$
        \State Compute scaled parameter update: $A_w^i = \Delta\theta_w^i \cdot \frac{\partial \mathcal{L}}{\partial \theta_w} \cdot f(\Delta \mathcal{L}^i)$
        \State Update sum of absolutes: $S_w = S_w + |A_w^i|$
        \State Store previous mean: $\mu_w^{prev} = \mu_w$
        \State Update running mean: $\mu_w = \mu_w + \frac{A_w^i - \mu_w}{i+1}$
        \State Update sum of squared differences: $SSD_w = SSD_w + (A_w^i - \mu_w^{prev}) \cdot (A_w^i - \mu_w)$
    \EndFor
\EndFor

\For{each weight $w$}
    \State Compute standard deviation: $\sigma_w = \sqrt{\frac{SSD_w}{N_w}}$
    \State Normalize sum of absolutes: $S_w^{norm} = \frac{S_w - \min(S)}{\max(S) - \min(S)}$
    \State Normalize standard deviation: $\sigma_w^{norm} = \frac{\sigma_w - \min(\sigma)}{\max(\sigma) - \min(\sigma)}$
    \State Compute preliminary profile: $P_w^{pre} = S_w^{norm} \cdot \sigma_w^{norm}$
    \State Compute final profile: $P_w = \frac{P_w^{pre} - \min(P^{pre})}{\max(P^{pre}) - \min(P^{pre})}$
\EndFor

\Return $P_w$ for all weights $w$
\end{algorithmic}
\end{algorithm}

\section{Experimental details}
\label{sec:experimental_details}
All model training runs are performed on NVIDIA A5000 GPUs, each with 24GB of memory and 309 TFLOPS of theoretical performance.
Specific training reproduce times cannot be provided as the experiments were conducted on an exclusive autonomous training platform that enables parallel execution and efficient management of multiple training runs.

\subsection{CIFAR10-CIFAR100 Setup Details}
\label{sec:cifar10_cifar100_setup_details}
We use a CNN architecture consisting of 4 convolutional layers followed by 2 dense layers with dropout. We create 6 separate heads of size 10 each, one for each task. The network is trained using our custom hook-enabled Adam optimizer with learning rate $\eta = 1 \times 10^{-3}$, $\beta_1 = 0.9$, $\beta_2 = 0.999$, and a minibatch size of 256. Each task is trained for 60 epochs.

For both CIFAR-10 and CIFAR-100 datasets, we use the standard split of 80\% for training and 20\% for validation, with equal distribution across classes. The training data undergoes random cropping with padding of 4 pixels, random horizontal flipping, conversion to tensor, and normalization using mean (0.5071, 0.4867, 0.4408) and standard deviation (0.2675, 0.2565, 0.2761). The validation data undergoes the same transformations except for the random augmentations (cropping and flipping).

We use a Gaussian hyperparameter $k=25$ and a regularization hyperparameter $r=0.03$.

\subsection{LOCO CIFAR100 Setup Details}
\label{sec:loco_cifar100_setup_details}
For the data split, we use 80\% of the data for training, and from the remaining 20\%, we randomly select 16\% for validation and 4\% for local testing. The training data undergoes random cropping with padding of 4 pixels, random horizontal flipping, conversion to tensor, and normalization using mean (0.4914, 0.4822, 0.4465) and standard deviation (0.2470, 0.2435, 0.2616). The validation and test data undergo the same transformations except for the random augmentations (cropping and flipping).

We use a Gaussian hyperparameter $k=25$ and a regularization hyperparameter of 0.05, 0.1, 0.2, and 0.3.
To provide a comprehensive comparison, we explored different regularization strengths for both SI and EWC. For SI, we tested K values of 0.05, 0.005, 0.0005, and 0.00005, and for EWC we used K values of 10, 100, 500, and 1000. These different K values control the regularization intensity, enabling us to examine various trade-offs between catastrophic forgetting mitigation and finetuning accuracy.

\section{Detailed Results}
\label{app:results}

\subsection{CIFAR10-CIFAR100 Results}
See Table~\ref{tab:cifar10_cifar100_results}.

The PPAP trainings were repeated 5 times to assess the statistical variance in accuracy.
\label{app:cifar10-cifar100}

\begin{table}[H]
\centering
\caption{CIFAR10-CIFAR100 Results}
\label{tab:cifar10_cifar100_results}
\begin{tabular}{lccccc}
\hline
Task & From Scratch & Fine-tuning & Consolidation (SI) & PPAP \\
\hline
1 & 0.853 & 0.427 & 0.812 & $0.820 \pm 0.006$ \\
2 & 0.732 & 0.488 & 0.680 & $0.736 \pm 0.012$ \\
3 & 0.696 & 0.498 & 0.702 & $0.740 \pm 0.014$ \\
4 & 0.780 & 0.455 & 0.786 & $0.751 \pm 0.025$ \\
5 & 0.688 & 0.686 & 0.747 & $0.760 \pm 0.009$ \\
6 & 0.777 & 0.832 & 0.801 & $0.823 \pm 0.008$ \\
\hline
\end{tabular}
\end{table}
\vspace{1em}

\subsection{LOCO CIFAR100 Results}
\label{app:loco-cifar100}
See Table~\ref{tab:loco_cifar100_results}.

\begin{table}[H]
\centering
\caption{LOCO CIFAR100 Results}
\label{tab:loco_cifar100_results}
\begin{tabular}{lcccccccc}
\hline
\multirow{2}{*}{Method} & \multicolumn{2}{c}{20-20} & \multicolumn{2}{c}{100-20} & \multicolumn{2}{c}{100-100} & \multicolumn{2}{c}{600-100} \\
& X & Y & X & Y & X & Y & X & Y \\
\hline
PPAP (r=0.1) & 0.751 & 0.804 & 0.793 & 0.796 & 0.779 & 0.813 & 0.803 & 0.805 \\
PPAP (r=0.05) & 0.764 & 0.790 & 0.801 & 0.771 & 0.787 & 0.804 & 0.809 & 0.793 \\
PPAP (r=0.3) & 0.720 & 0.812 & 0.771 & 0.817 & 0.761 & 0.825 & 0.790 & 0.821 \\
PPAP (r=0.2) & 0.733 & 0.810 & 0.781 & 0.811 & 0.769 & 0.820 & 0.795 & 0.815 \\
SI (rc=0.05) & 0.709 & 0.777 & 0.789 & 0.793 & 0.788 & 0.782 & 0.811 & 0.771 \\
SI (rc=0.005) & 0.691 & 0.800 & 0.769 & 0.806 & 0.768 & 0.813 & 0.799 & 0.795 \\
SI (rc=0.0005) & 0.670 & 0.809 & 0.749 & 0.823 & 0.752 & 0.830 & 0.788 & 0.814 \\
SI (rc=0.00005) & 0.666 & 0.802 & 0.735 & 0.823 & 0.724 & 0.834 & 0.779 & 0.824 \\
EWC (rc=10) & 0.721 & 0.799 & 0.755 & 0.829 & 0.761 & 0.828 & 0.756 & 0.832 \\
EWC (rc=100) & 0.737 & 0.780 & 0.710 & 0.810 & 0.712 & 0.799 & 0.713 & 0.834 \\
EWC (rc=500) & 0.764 & 0.765 & 0.782 & 0.790 & 0.788 & 0.775 & 0.757 & 0.830 \\
EWC (rc=1000) & 0.737 & 0.766 & 0.710 & 0.783 & 0.712 & 0.770 & 0.713 & 0.830 \\
\hline
\end{tabular}
\end{table}

\end{document}